\documentclass{article}
\usepackage{ijcai18}

\usepackage{times}
\usepackage{xcolor}
\usepackage[utf8]{inputenc}
\usepackage[small]{caption}

\usepackage{graphicx}
\usepackage{lineno}
\usepackage{times}
\usepackage{helvet}
\usepackage{courier}
\usepackage{epsfig}
\usepackage{graphicx}
\usepackage{subfigure}
\usepackage{amsmath}
\usepackage{array}
\usepackage{amssymb}
\usepackage{multirow}
\usepackage{epstopdf}
\usepackage{rotating}
\usepackage{CJK}
\usepackage{booktabs}
\usepackage{color}
\usepackage{graphicx}

\newcommand{\eat}[1]{}
\title{From Pixels to Objects: Cubic Visual Attention for Visual Question Answering}

\author{Jingkuan Song, Pengpeng Zeng, Lianli Gao, Heng Tao Shen\thanks{Corresponding author: Lianli Gao, Heng Tao Shen}\\
	Center for Future Media and School of Computer Science and Engineering, \\University of Electronic Science and Technology of China, Chengdu 611731, China.\\ 
	\{lianli.gao, pengpeng.zeng\}@uestc.edu.cn, {jingkuan.song}@gmail.com and shenhengtao@hotmail.com}
\begin{document}

\maketitle

\begin{abstract}
%Question-guided Visual attention mechanisms are well studied by utilizing question semantic representation as a query to selectively target different visual areas that are related to the answer.
%Those attention attentions have been successfully applied in Visual question answering (VQA) to address "where to look" problem by directly utilizing a high-level conv feature map. A high-level filter/channel within a deep Convolutional Neural Network (CNN) has potential to detect rich evidence of objects parts or complete objects or semantic attributes. However, existing VQA approaches ignore the importance of channel-based information. In this paper we propose a stacked visual attention (CVA) framework by successively applying a novel channel attention and spatial or region attention to improve VQA task. In addition, it explores the suitability for VQA of region-wise representations pooled from an object detection CNN such as Faster R-CNN. We take advantage of the object proposals learned by a Region Proposal Network (RPN) and their associated conv features to build a VQA pipeline composed of a first stacked visual attention followed by simply question-guided attention. We assess the performance of our proposed CVA on three public image QA datasets, including COCO-QA, VQA and Visual7W. Experimental results show that our proposed method significantly outperforms the state-of-the-arts.    

Recently, attention-based Visual Question Answering (VQA) has achieved great success by utilizing question to selectively target different visual areas that are related to the answer.
Existing visual attention models are generally planar, i.e., different channels of the last conv-layer feature map of an image share the same weight. This conflicts with the attention mechanism because CNN features are naturally spatial and channel-wise. Also, visual attention models are usually conducted on pixel-level, which may cause region discontinuous problem.
In this paper we propose a Cubic Visual Attention (CVA) model by successfully applying a novel channel and spatial attention on object regions to improve VQA task. 
Specifically, instead of attending to pixels, we first take advantage of the object proposal networks to generate a set of object candidates and extract their associated conv features. Then, we utilize the question to guide channel attention and spatial attention calculation based on the con-layer feature map. Finally, the attended visual features and the question are combined to infer the answer. We assess the performance of our proposed CVA on three public image QA datasets, including COCO-QA, VQA and Visual7W. Experimental results show that our proposed method significantly outperforms the state-of-the-arts.

\end{abstract}

\section{Introduction}

Visual Question Answering (VQA) is an interdisciplinary research problem, which has attracted extensive attention recently \cite{YangHGDS15,Lu2016Hie,dual2016,lu2018:co-attending}. It has the potential to be applied for assisting the visually impaired people and automatically querying on large-scale image or video datasets \cite{xu2015ask,dual2016}. Compared with image captioning task, the VQA task requires a deeper understanding of both image and question rather than a coarse understanding \cite{DBLP:conf/cvpr/GoyalKSBP17,vqa2015}. It inspects intelligent system's ability by inferring a correct answer for the visual question.
\begin{figure}[t]
	\centering\includegraphics[width=1\linewidth]{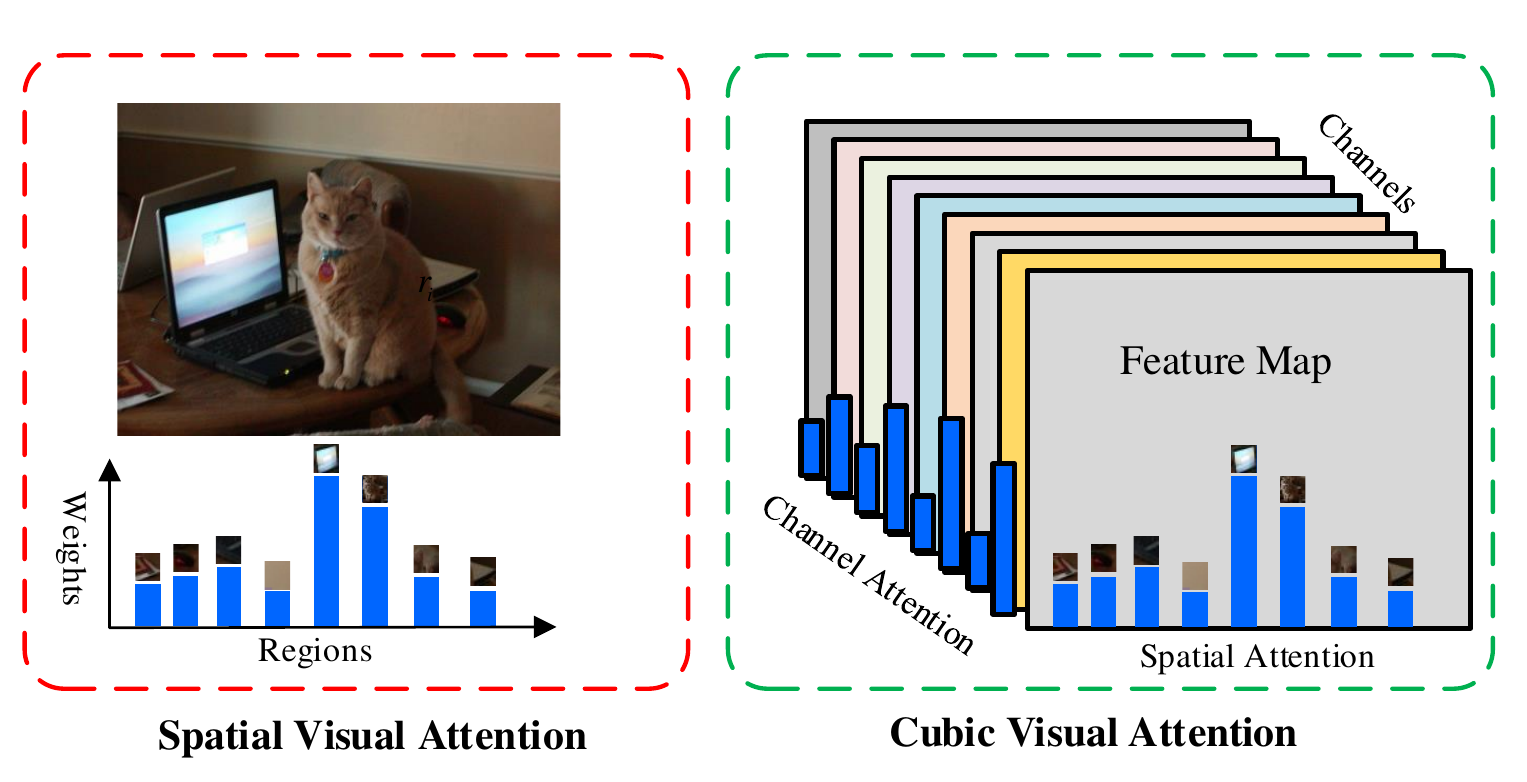}
	\caption{The illustration of spatial visual attention and cubic visual attention. Spatial visual attention is modelled as a weight matrix on the last conv-layer feature map of a CNN encoding an input image. Different channels share the same weights. Instead, cubic visual attention learns both spatial and channel attention, which conforms to the nature of conv features, i.e., spatial and channel-wise.}
	\label{fig:0}
\end{figure}

\begin{figure*}[t]
	\centering\includegraphics[width=0.92\linewidth]{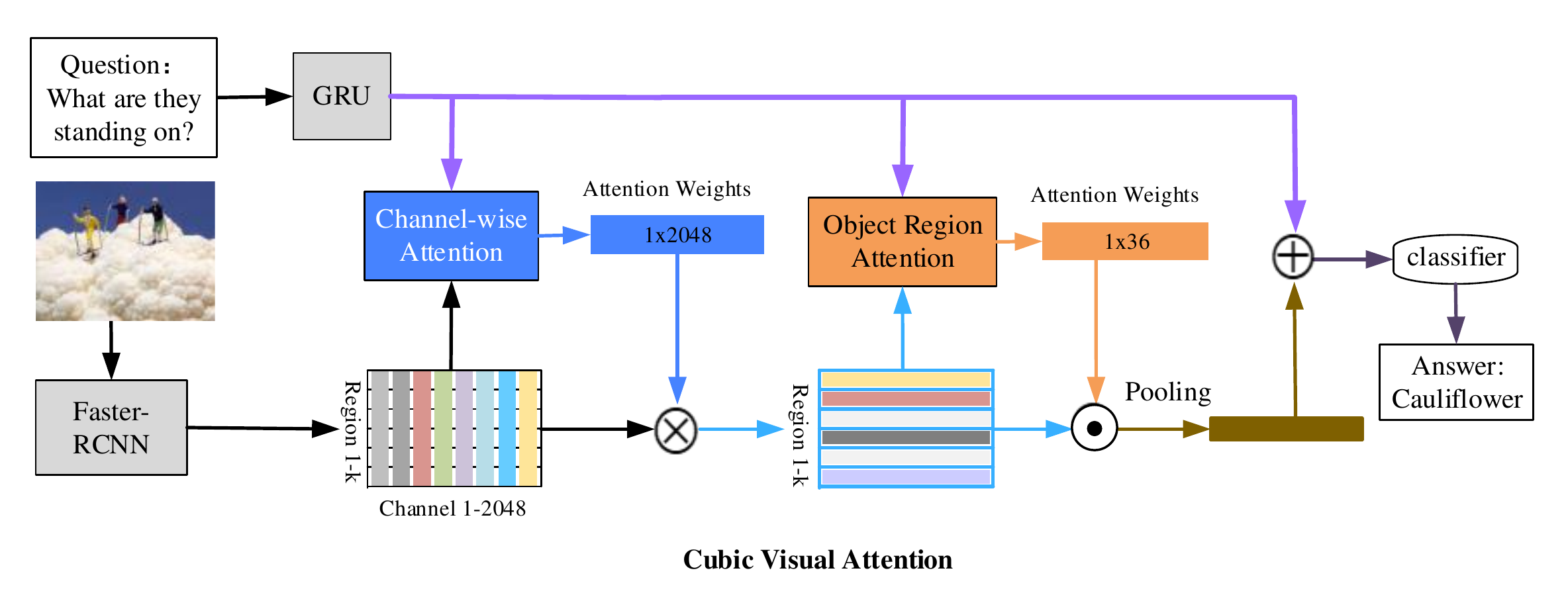}
	\caption{The framework of our proposed CVA for VQA. It consists of four main components, feature extraction, object proposal, cubic visual attention estimation and answer prediction.}
	\label{fig:1}
\end{figure*}

Existing works for VQA can be generally classified into two categories: 1) Typical CNN-RNN models, which transfer image captioning frameworks by integrating CNN with Recurrent Neural Networks (RNN) to solve VQA tasks \cite{gao2015you,nips2015,DBLP:conf/iccv/MalinowskiRF15}; and 2) Question-guided visual attention mechanisms, which aim to discover the most important regions to answer a question by exploring their relationships \cite{DBLP:conf/emnlp/FukuiPYRDR16,Lu2016Hie,dual2016,DBLP:conf/cvpr/ShihSH16,xu2015ask,YangHGDS15}. More specifically, question-guided visual attentions are conducted by concatenating the semantic representation of a question with each candidate region and then put them into a multiple layer perceptron (MLP) or applying the dot product of each word embedding and each spatial location's visual feature \cite{xu2015ask}. In addition, Yang \textit{et al.} \cite{YangHGDS15} proposed a stacked attention model by utilizing semantic representation of a question as query to search for the regions in an image multiple times to infer an answer progressively. In \cite{dual2016}, a dual attention was introduced to infer the answers by attending to specific regions in images and words in text. Lu \textit{et al.} \cite{Lu2016Hie} presented a co-attention that jointly performs question-guided visual attention and image-guided question attention to address `where to look' and `what words to listen to'. 

Although promising results have been achieved, typical CNN-RNN models resort to a global image presentation which may contain noisy or unnecessary information for the related question. To some extent, question-guided visual attention mechanisms have tackled this problem generally by spatial attention, i.e., the attention is modeled as spatial probabilities that re-weight the last conv-layer feature map of a CNN encoding an input image. However, such spatial attention does not necessarily conform to the attention mechanism--a dynamic feature extractor that combines contextual fixations over time, as CNN features are naturally spatial and channel-wise \cite{chen2016sca}.
Image features are generally extracted by deep Convolutional Neural Networks (CNNs) \cite{HeZRS16,SongGGLZS17,GaoGZXS17}. Starting from an input color image of the size $W \times H \times3$, the last convolutional layer consisting of $C$-channel filters output a $W' \times H' \times C$ feature map. 
Different channels of a feature map is essentially activation response maps of the corresponding filter, and channel-wise attention can be viewed as the process of selecting semantic attributes on the demand of the sentence context.  For example, when we want to predict \textit{cat}, our channel attention (e.g., in the conv5 3/conv5 4 feature map) will assign more weights on channel-wise feature maps generated by filters according to the semantics like furry texture, ear, and cat-like shapes. Channel attention plays a different role compared with spatial attention, and it is rarely addressed in previous works.

%In CNN, a high-level filter channel is able to detect rich evidence of objects parts or complete objects or semantic attributes, which could provide additional supplementary information for predicting an answer.  However, previous attention-based models calculate attention weights based on the  semantic similarity between the question and image regions, i.e., those models merely moderate sentence semantics into the last conv-layer via visually spatial attention weights. The cubic attention is rarely addressed.

In this paper, we take the full advantage of two characteristics (i.e., channel and spatial) of object-based region features for visual attention-based VQA. Specifically, we propose a novel Cubic Visual Attention (CVA) framework by successfully applying a channel attention and a spatial attention to assist VQA task. An object detection network \cite{DBLP:journals/pami/RenHG017}, which has the potential to enable nearly cost-free object region proposal, is applied to extract top-k objects in an image and each object is represented by a D-dimensional vector. Next, the channel-wise attention learns to pay attention to specific channels of the last conv feature map. Thirdly, a region-based spatial attention is applied on the channel-attended features to select related objects. Finally, an answer is inferred by considering both the attended visual features and the question. %We validate the effectiveness of the proposed CVA framework on three well-known image question answering benchmarks, including COCO-QA, VQA and Visual7W. Experimental results show that our proposed CVA significantly surpass the state-of-the-art. 

\section{Cubic Visual Attention for VQA}

The VQA task is to predict an answer from a question and a related image. 
In this section, we introduce our proposed CVA framework (shown in Fig.~\ref{fig:1}) for VQA, and it consists of 1) a feature extraction component, which extracts the features for question and input image; 2) a channel attention for selecting filters related to the high-level object semantic; 3) a region-based spatial attention for learning to focus on the regions of the image that are important and 4) an answer prediction layer to infer to answer. In this section, we describe each of them.

\subsection{Input Representations}
\textbf{Image Features.} 
For the input image, we use an existing state-of-the-art object detection network Faster R-CNN \cite{DBLP:journals/pami/RenHG017} for object proposal and feature extraction. Specifically, each image is input into Faster R-CNN model to obtain the top-$K$ candidate objects. For each selected region/box $k$, ${\textbf{v}_k}$ is defined as the mean-pooled convolutional feature from this region and each ${\textbf{v}_k}$ indicates a vector with $D$ dimensions. Therefore, the input image $\textbf{V}$ is defined as follow:
\begin{eqnarray}
\textbf{V} = [\textbf{v}_{1}, \textbf{v}_{2}, ..., \textbf{v}_{K}], \textbf{v}_{k} \in \mathbb{R}^{1\times D}
\end{eqnarray}

%\textcolor{red}{write something about R-CNN}

\noindent\textbf{Encoding Question Features.}
Suppose that $ [ \textbf{q}_{1}, \textbf{q}_{2}, ..., \textbf{q}_{T} ] $ represents an question, where \( \textbf{q}_{t} \) is an one-hot representation for the word at position $t$, and $T$ is the length of the question. Each word representation \(\textbf{q}_{t} \) is transferred into a lower dimensional vector \( \textbf{x}_{t} \) with an embedding matrix \( \textbf{W}_{e}^{\textbf{q}}  \).
\begin{eqnarray}
\textbf{x}_{t}&=&\textbf{W}_{e}^{q}\textbf{q}_{t}
\end{eqnarray}
There are various approach to encode a question like bag-of-words. However, Long Short-term Memory (LSTM) and gated recurrent unit (GRU) are two of the most popular mechanisms to encode a sentence in machine translation and great results have been obtained. In this paper, we employ GRU to encode our questions, and the gradient chains of GRU do not vanish due to the length of questions. For the $t$-th time step, the GRU unit takes the embedding vector \( \textbf{x}_{t} \) as an input, updates the gate \( \textbf{z}_{t} \), resets gate \( \textbf{r}_{t} \), and then outputs a hidden state \(\textbf{h}_{t}\). After $T$ steps, we obtain the $T$-th output \(\textbf{h}_{T}\) to represent the semantic information of a question. We formulate our encoding process as below:
\begin{eqnarray}
\textbf{z}_{t}&=& \sigma(\textbf{W}_{z}\textbf{x}_{t} + \textbf{U}_{z}\textbf{h}_{t-1} + \textbf{b}_{z})\\
\textbf{r}_{t}&=& \sigma(\textbf{W}_{r}\textbf{x}_{t} + \textbf{U}_{r}\textbf{h}_{t-1} + \textbf{b}_{r})\\
\tilde{\textbf{h}}_{t}&=& tanh(\textbf{W}_{h}\textbf{x}_{t} + \textbf{U}_{h}(\textbf{r}_{t} \circ \textbf{h}_{t-1}) + \textbf{b}_{h})\\
\textbf{h}_{t}&=& \textbf{z}_{t} \circ \textbf{h}_{t-1} + (1 - \textbf{z}_{t}) \circ \tilde{\textbf{h}}_{t} \\
\textbf{Q}&=&\textbf{h}_{T}
\end{eqnarray}
where \( \textbf{W}_{z,r,h}, \textbf{U}_{z,r,h}, \textbf{b}_{z,r,h} \) are the parameter which needed to be learn. Note that \(\sigma\) is a sigmiod activation function, and \(\circ\) is used as the Hadamard product or element-wise multiplication. Through the question feature encoding process, we generate the question representation $\textbf{Q}$, which equals to the last output of GRU ($\textbf{h}_{T}$).

\subsection{Channel Attention}
In this section, we introduce a novel channel-wise attention mechanism to attend the visual features $\textbf{V}$. Each $\textbf{v}_{i}$ is obtained by mean pooling the conv features of a given box spatial location. Essentially, each channel of a feature map in CNN is correlated to a convolutional filter which performs as a pattern detector. For instance, the lower-level filters detect visual clues such as edges and color, while the higher-level filters detect semantic patterns, such as attributes or object components. In this work, our object region features are pooled from the last conv feature map, thus each channel represents the semantic patterns of the detected objects within an image. Therefore, conducting a channel-wise attention can be viewed as a process of choosing object semantic attributes.

For channel-wise attention, we first reshape $\textbf{V}$ to $\textbf{U}$ and \( \textbf{U} = [\textbf{u}_{1}, \textbf{u}_{2}, ..., \textbf{u}_{D}] \), where \( \textbf{u}_{i} \in \mathbb{R}^k\) represents the i-th dimension of the whole object feature $V$, and $D$ is the dimension of $\textbf{v}_{i}$ or it is the total number of channels for each object region. Next, we apply a mean pooling for each channel to generate the channel vector 
\begin{eqnarray}
\overline {\textbf{u}} = [\overline {{u}}_{1}, \overline {{u}}_{2}, ..., \overline {{u}_{D}}] 
\end{eqnarray}
where $\overline {{u}_{i}}$ is the mean vector of $\textbf{u}_{i}$, which represents the $i$-the channel features. Our channel-wise attention ${{\rm \textbf{A}}_c}$ is defined as below:
\begin{eqnarray}
\textbf{b} &=& tanh( (\textbf{W}_{{vc}}\overline {\textbf{u}} + \textbf{b}_{vc}  ) \otimes (\textbf{W}_{qc}\textbf{Q} + \textbf{b}_{qc}  ) ) \\
\beta &=& softmax( \textbf{W}_{c}\textbf{b} + \textbf{b}_{c} )
\end{eqnarray}
where $\textbf{W}_{vc}$, $\textbf{W}_{qc}$ and $ \textbf{W}_{c}$ are embedding matrices, $\textbf{b}_{vc}$, $\textbf{b}_{qc}$ and $ \textbf{b}_{c}$ are bias terms, and \( \otimes \) indicates the outer product of vectors. To sum up, we obtain our channel-wise attention weight $\beta$ through our channel-wise attention ${{\rm \textbf{A}}_c}$, defined as follow:
\begin{eqnarray}
\beta  = {{\rm \textbf{A}}_c}\left( {\overline{ \textbf{U}} ,\textbf{Q}} \right)
\end{eqnarray}

\subsection{Object Region-based Spatial Attention}
With above step, we obtain the channel-wise attention weight $\beta$, thus we can feed $\beta$ to a channel-wise attention function ${\textbf{f}_c}$ to calculate a modulated feature map ${\textbf{V}^c}$:
\begin{eqnarray}
{\textbf{V}^c} = {\textbf{f}_c}\left( {\beta ,\textbf{V} } \right)
\end{eqnarray}
where ${\textbf{f}_c}$ is an channel-wise multiplication for region feature map channels and corresponding channel weights. In addition, eventually ${\textbf{V}^c}$ is
\begin{eqnarray}
{\textbf{V}^c} = \left\{ {\textbf{v}_1^c,\textbf{v}_2^c,...,\textbf{v}_k^c} \right\},\textbf{v}_i^c \in {\mathbb{R}^D}
\end{eqnarray}

Suppose, we have ${\textbf{V}^c}$ where $\textbf{v}_i^c$ indicates the visual feature of the $i$-th object region. In general, a question may only relate to one or several particular regions of an image. If we want to ask `what the color of the dog', then only the dog object region contains the useful information, therefore typical CNN-RNN which employing the whole global visual feature may lead to sub-optimal results due to the irrelevant visual regions shown in the input image. Instead of considering each object region equally, our region-based spatial attention mechanism aims to target the most related region with an referred question. Given the previous calculated ${\textbf{V}^c}$, a single-layer neural network is adopted to take both ${\textbf{V}^c}$ and $\textbf{Q}$ as inputs to generate a new feature $\textbf{a}$, and then a softmax function is followed to compute the region-based spatial attention weight $\eta$. The object region-based spatial attention ${{\rm \textbf{A}}_s}$ is defined as :
\begin{eqnarray}
\textbf{a} &=& tanh( \textbf{W}_{vo}\textbf{V}^c + \textbf{b}_{vo}) \oplus ( \textbf{W}_{qo}Q + \textbf{b}_{qo}) \\
\eta &=& softmax( \textbf{W}_{o}\textbf{a} + \textbf{b}_{o} )
\end{eqnarray}
where $\textbf{W}_{vo}$ and $\textbf{W}_{qo}$ are the embedding matrices that project both visual and question features into a common latent space. In addition, $\textbf{W}_{o}$ is a set of parameters that needs to be learn. $\textbf{b}$ is the model bias. \( \oplus \) is the addition of a matrix and a vector. Moreover, \(\eta \in \mathbb{R}^k \) is a $k$-dimensional vector, which represents the importance of each object region. Therefore, the weights $\eta$ can be calculated with the following simple presentation:
\begin{eqnarray}
\eta  = {{\rm \textbf{A}}_s}\left( {{\textbf{V}^c},\textbf{Q}} \right)
\end{eqnarray}
Furthermore, to deal with multiple object regions, a simple strategy usually is used to compute the average of features across the whole image, and this generated feature is used as input to integrate question feature to generate an answer:
\begin{eqnarray}
{\textbf{V}^s} = \frac{1}{k}\sum\limits_{i = 0}^k {\textbf{v}_i^c} 
\end{eqnarray}
However, as mentioned above this strategy effectively effectively integrate multiple regions into a single vector, neglecting the inherent spatial structure and leading to the loss of information. Instead of using above simple strategy, we apply $\eta$ to attend where to look at and it defined as bellow:
\begin{eqnarray}
{\textbf{V}^s} = {\textbf{f}_s}\left( {\eta ,{\textbf{V}^c}} \right) = \frac{1}{k}\sum\limits_{i = 0}^k {{\eta _i}\textbf{v}_i^c} 
\end{eqnarray}

\subsection{Answer Prediction}
Following previous work \cite{Wang2017fact,Lu2016Hie}, we treat the answer prediction process as a  multi-class classification problem, in which each class corresponds to a distinct answer. We predict the answer based on the stacked attended image visual feature ${\textbf{V}^s}$ and a question feature $\textbf{Q}$, and a multi-layer perceptron (MLP) is used for classification:
\begin{eqnarray}
h &=& tanh(\textbf{W}_{v}\textbf{V}^s + \textbf{W}_{q}\textbf{Q} + \textbf{b}_{h}) \\
p &=& softmax( \textbf{W}_{h}  \textbf{h} + \textbf{b}_{p} )
\end{eqnarray}
where \( \textbf{W}_{v}, \textbf{W}_{q}, \textbf{W}_{h} \) are parameters. \( \textbf{b}_{h}, \textbf{b}_{p} \) are bias terms and $\textbf{p}$ is the probability of the final answer.

\subsection{A Variant of CVA}
The previous introduced stacked attention mechanism applies channel-wise attention before spatial attention. Given an initial question feature $\textbf{Q}$ and visual region features $\textbf{V}$, we adopt a channel-wise attention ${{\rm \textbf{A}}_c}$ to compute the channel-wise attention weights $\beta$ for obtaining a channel-wised weighted object region features ${\textbf{V}^c}$. Next, we apply the object region based spatial attention ${{\rm \textbf{A}}_s}$ by taking ${\textbf{V}^c}$ as inputs to obtain region spatial weights $\eta$. The pipeline of this framework can be summarized as follows:
\begin{eqnarray}
\beta  = {{\rm \textbf{A}}_c}\left( {\overline {\textbf{U}} ,\textbf{Q}} \right)\\
{\textbf{V}^c} = {\textbf{f}_c}\left( {\beta ,\overline {\textbf{U}} } \right)\\
\eta  = {{\rm \textbf{A}}_s}\left( {{\textbf{V}^c},\textbf{Q}} \right)\\
{\textbf{V}^s} = {\textbf{f}_s}\left( {\eta ,{\textbf{V}^c}} \right)
\end{eqnarray}

In order to further study the effect of the oder of channel-wise and spatial attentions, we propose an CVA variant which exchange the order of two attentions by firstly applying spatial attention  ${{\rm \textbf{A}}_s}$ and then following by a channel-wise attention ${{\rm \textbf{A}}_c}$. This pipeline can be summarized as follows:
\begin{eqnarray}
\eta  = {{\rm \textbf{A}}_s}\left( {\textbf{V},\textbf{Q}} \right)\\
{\textbf{V}^s} = {\textbf{f}_s}\left( {\eta ,\textbf{V}} \right)\\
\beta  = {{\rm \textbf{A}}_c}\left( {{\textbf{U}^s},\textbf{Q}} \right)\\
{\textbf{V}^c} = {\textbf{f}_c}\left( {\beta ,{\textbf{V}^s}} \right)
\end{eqnarray}
where ${\textbf{U}^s}$ is obtained by reshape ${\textbf{V}^s} $, seen channel-wise reshape operation. Further more, ${\textbf{V}^c}$ and $\textbf{Q}$ are utilized to predict the final answer.  For simplicity, we name this pipeline as CVA-V.

\section{Experiments}
\subsection{Datasets}
We evaluate our proposed model on three public image QA datasets: the COCO-QA dataset, the VQA dataset and Visual7W dataset. 

\textbf{COCO-QA dataset.}  This dataset \cite{nips2015} is proposed to enable training large complex models due to the reason that the previous DAQUAR dataset only contains approximately 1,500 images and 7,000 questions on 37 common object classes. Therefore, COCO-QA is created to produce a much larger number of QA pairs and a more evenly distributed answers based on the MS-COCO dataset. It contains in total 117,684 samples with 78,736 as training and 38,948 as testing. There are 23.29\% and 18.7\% overlap in training questions and training question-answer pairs. COCO-QA consists of four question categories: \textit{Object} (54,992 Training vs 27,206 Testing), \textit{Number} (5,885 vs 2,755), \textit{Color} (13,059 vs 6,509) and \textit{Location} (4,800 vs 2,478). In addition, all answers in this dataset are a single word.

\textbf{VQA dataset.} It is a large-scale dataset presenting both open-ended answering task requiring a free-form response and multiple-choice task requiring an approach to pick from a predefined list of possible answers. Specifically, 204,721 real images (123,287 training and validation vs 81,434 testing ) are collected from the newly-released Microsoft Common Objects in Context (MS COCO) dataset. For this VQA dataset challenge, it contains two test categories: 1) test-dev for debugging and validation purpose; and 2) test standard indicates the `standard' test data for the VQA competition. More specifically, we use 248,349 training questions, 121,512 validation questions and 244,302 test questions. The question types include \textit{Y/N} 38.37\%, \textit{Number} 12.31\% and \textit{Other}. Compared with COCO-QA, the answer form is diversity with an answer containing one, two or three words respectively being 89.32\%, 6.91\%, and 2.74\%. In addition, for each image, three questions are collected and each question is answered by ten subjects alone with confidence. Following ~\cite{dual2016}, we set th number of possible answer for VQA as 2,000.

\textbf{Visual7W.} This dataset is collected recently by Zhu \textit{et al} \cite{DBLP:conf/cvpr/ZhuGBF16} with 327,939 QA pairs on 47,300 COCO images, which is a subset of Visual Genome image dataset. With the Amazon Mechanical Turk (AMT), they collected 1,311,756 human-generated multiple-choices and 561,459 object groundings from 36,579 categories.  For each QA pair, it has four human-generated multiple-choices and only one of them is correct. In addition, for Visual7W, it consists of seven types of question including \textit{what},\textit{where},\textit{when}, \textit{who},\textit{why},\textit{how} and \textit{which}. The first six type of questions are proposed to examine the capability of a model of visual understanding. Compared with VQA, Visual7W contains rich questions and longer answers. Moreover, Visual7W establishes an explicit link between QA pairs and image regions by providing complete grounding annotations and providing diverse question to acquire detailed visual information. Following \cite{mutil2017attention}, we only test our model in the settings of multiple-choices.

\begin{table}[t]
	\centering
	\begin{tabular}{|l|llll|}
		\hline
		Methods             & Y/N   & Num.  & Other & All   \\ \hline \hline
		CA & 80.55 & 39.10 & 52.45 & 62.54 \\
		RA    & 83.41 & 39.01 & 55.91 & 65.37 \\
		CVA   & \textbf{83.73} & \textbf{40.91}  & \textbf{56.36}  & \textbf{65.92} \\ 
		R-CVA    & 83.39 & 40.89 & 55.86 & 65.54 \\ \hline
	\end{tabular}
	\caption{Ablation study on the VQA test-dev dataset}
	\label{TAB1}
\end{table}

\begin{table*}[t]
	\centering
	\small
	\begin{tabular}{|c||cccccccccc|}
		\hline
		\multirow{3}{*}{Approach} & \multicolumn{5}{c|}{test-dev}                               & \multicolumn{5}{c|}{test-standard}      \\ \cline{2-11} 
		& \multicolumn{4}{c|}{Open Ended} & \multicolumn{1}{c|}{MC}   & \multicolumn{4}{c|}{Open Ended} & MC    \\ \cline{2-11} 
		& Y/N    & Num    & Other & All   & \multicolumn{1}{c|}{All}  & Y/N    & Num    & Other & All   & All   \\ \hline\hline
		LSTM Q +I                 & 78.9   & 35.2   & 36.4  & 53.7  & \multicolumn{1}{c|}{57.2} & 79.0   & 35.6   & 36.8  & 54.1  & 57.8  \\
		deeper +norm              & 80.5   & 36.8   & 43.1  & 57.8  & \multicolumn{1}{c|}{62.7} & 80.6   & 36.5   & 43.7  & 58.2  & 63.1  \\
		DPPnet                    & 80.7   & 37.2   & 41.7  & 57.2  & \multicolumn{1}{c|}{-}    & -      & -      & -     & 58.9  & -     \\ \hline
		
		SAN                       & 79.3   & 36.6   & 46.1  & 58.7  & \multicolumn{1}{c|}{-}    & -      & -      & -     & 59.5  & -     \\
		FDA                       & 81.1   & 36.2   & 45.8  & 59.2  & \multicolumn{1}{c|}{-}    & -      & -      & -     & 60.4  & -     \\
		DMN+                      & 80.5   & 36.8   & 48.3  & 60.3  & \multicolumn{1}{c|}{-}    & -      & -      & -     & -     & -     \\
		MCB                       & 81.2   & 35.1   & 49.3  & 60.8 & \multicolumn{1}{c|}{65.4} & -      & -      & -     & -     & -     \\
		\eat{	MCB+Att                   & 82.2   & 37.7   & 54.8  & 64.2  & \multicolumn{1}{c|}{68.6} & -      & -      & -     & -     & -     \\
			MCB+Att.+GloVe            & 82.5   & 37.6   & 55.6  & 64.7  & \multicolumn{1}{c|}{69.1} & -      & -      & -     & -     & -     \\
			MCB+Att.+GloVe+VG            & 82.3   & 37.2   & \textbf{57.4}  & 65.4  & \multicolumn{1}{c|}{69.9} & -      & -      & -     & -     & -     \\}
		\textcolor{gray}{MCB-7}    & \textcolor{gray}{83.4}   & \textcolor{gray}{39.8}  & \textcolor{gray}{58.5}  & \textcolor{gray}{66.7}  & \multicolumn{1}{c|}{\textcolor{gray}{70.2}} & \textcolor{gray}{83.2}     & \textcolor{gray}{39.5}     & \textcolor{gray}{58.0}     & \textcolor{gray}{66.5}     & \textcolor{gray}{70.1}    \\ 
		\textit{Our CVA }                 & \textbf{83.73}  & \textbf{40.91}  & \textbf{56.36} & \textbf{65.92} & \multicolumn{1}{c|}{\textbf{70.3}} & \textbf{83.79}  & \textbf{40.41}  & \textbf{56.77} & \textbf{66.20} & \textbf{70.41} \\ \hline

		AC                        & 79.8   & 36.8   & 43.1  & 57.5  & \multicolumn{1}{c|}{-}    & 79.7   & 36.0   & 43.4  & 57.6  & -     \\
		ACK                       & 81.0   & 38.4   & 45.2  & 59.2  & \multicolumn{1}{c|}{-}    & 81.1   & 37.1   & 45.8  & 59.4  & -     \\ \hline
		HieCoAtt                  & 79.7   & 38.7   & 51.7  & 61.8  & \multicolumn{1}{c|}{65.8} & -      & -      & -     & 62.1  & 66.1  \\
		DAN                       & 83.0   & 39.1   & 53.9  & 64.3  & \multicolumn{1}{c|}{69.1} & 82.8   & 39.1   & 54.0  & 64.2  & 69.0  \\ \hline
		MLAN            & 82.9   & 39.2   & 52.8  & 63.7  & \multicolumn{1}{c|}{68.9} & -      & -      & -     & -     & -     \\ \hline
		
	\end{tabular}
	\caption{Comparison results on VQA dataset. According to different attention mechanisms, all the approaches are divided into five categories and each row represents one category. Row One indicates \textit{No Attention}. Row Two utilizes \textit{Visual Attention only}. Row Three applies \textit{Semantic Attention}. Row Four includes both \textit{Visual and Question Attentions}. Row Five applies both \textit{Semantic and Visual Attentions}. }
	\label{TAB2}
\end{table*}

\subsection{Evaluation Metrics}
The VQA task is usually regarded as a multi-class classification problem, and thus accuracy is an important evaluation metric for evaluating the performance of VQA models. Following \cite{vqa2015}, we use the following equation to compute the classification accuracy:
\begin{eqnarray}
Acc(ans) = min \left  \{ \frac{\#humans \ that \ said \ ans}{3}, 1 \right \} 
\end{eqnarray}
where $ans$ is the answer predicted by a VQA model. 

In addition, for MS COCO dataset, we also report the performance in terms of the Wu-Palmer similarity (WUPS), which accounts for word-level ambiguities in the answer words. The equation is defined as \cite{DBLP:conf/iccv/MalinowskiRF15} and it contains a thresholded taxonomy-based Wu-Palmer similarity parameter. For COCO-QA, we report WUPS at two extremes, 0.0 and 0.9.

\subsection{Implementation Details}
For extracting visual object features, we first integrate Faster R-CNN with ResNet-101 retrained on the ImageNet dataset by following an image captioning approach \cite{Anderson:2017up-down} and then select top 36 ($k=36$) object regions and each region is represented as 2,048 dimensional features. For sentence encoding, a pre-trained GloVe word embedding of dimension (300) and a single layer GUR are utilized. In addition, the dimension of every hidden layer including GRU, attention models and the final joint feature embedding is set as 1,024.

In our experiments, our models are trained with Adam. The batch size is set to 256, and the epoch is set as 30. More specifically, gradient clipping technology and dropout are exploited in training.

\subsection{Ablation Study}
For VQA challenge, it contains the test-dev, which is proposed to debug and validate VQA models, thus VQA competition evaluation server allows for unlimited submission. In this section, we perform ablation study on the VQA dataset to qualify the role of each component in our model. Specifically, we re-train our approach by ablating certain components: 1) channel-wise attention only (CA); 2) object region attention only (RA); 3) our stacked attention with both channel-wise and region-based attention (CVA); and 4) reversed stacked attention (R-CVA) by changing the order of channel-wise and region-based attention to test whether their order effect the VQA performance.

The experimental results are shown in Tab.\ref{TAB1} (test-standard is not recommended to be used for VQA ablation study). From the experimental results, we can see that object region attention alone performs better than channel-wise attention only on the \textit{Y/N}, \textit{Other} and \textit{All} with an increase of 2.86\%, 3.46\% and 2.83\% respectively. In terms of \textit{Number}, channel-wise attention performs slightly better. By stacking those two attention into a VQA model in any order, we find that the stacked attention models improve the performance of VQA, especially for \textit{Number} by approximately 1.9\%. In addition, changing order would slightly effect the performance.

\begin{figure*}[t]
	\centering\includegraphics[width=0.8\linewidth]{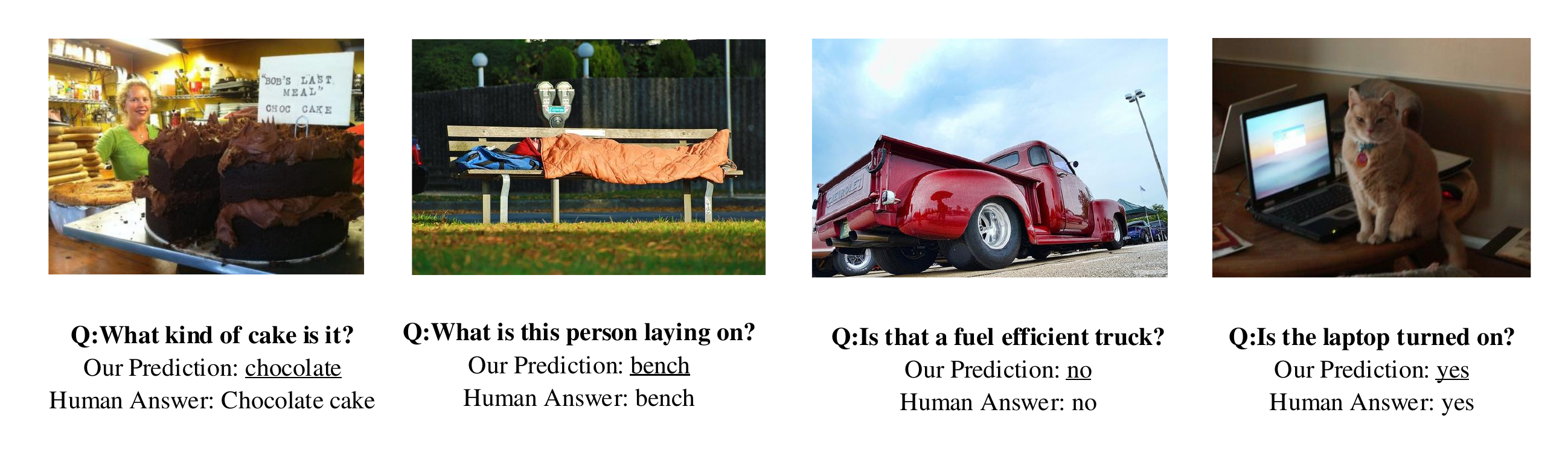}
	\caption{Four qualitative results from visual question answering.}\label{fig:2}
\end{figure*}

\subsection{Comparing with State-of-the-Arts on the VQA dataset}
\textbf{Compared Methods.} We compare our opposed CVA with the state-of-the-art VQA approaches, which can be divided into four categories:1) No attention approaches (LSTM Q+I \cite{vqa2015} and deeper+norm \cite{vqa2015} and DPPnet \cite{DBLP:conf/cvpr/NohSH16}); 2) visual attention based methods (SAN \cite{YangHGDS15}, FDA \cite{DBLP:journals/corr/IlievskiYF16}, DMN+ \cite{DBLP:conf/icml/XiongMS16}, MCB+Att. \cite{DBLP:conf/emnlp/FukuiPYRDR16} and MCB-7 (ensemble of 7 Att. models) \cite{DBLP:conf/emnlp/FukuiPYRDR16}); 3) Utilizing high-level concepts as visual features or semantic attention based methods (AC \cite{ask16anything} and ACK \cite{ask16anything} );  4) methods with both image attention and question attention (HieCoAtt \cite{Lu2016Hie} and DAN \cite{dual2016}) and 5) jointly learning semantic attention and visual attention (MLAN (ResNet) \cite{mutil2017attention}). Our proposed method CVA belongs to the second category, visual attention only.

\textbf{Results.} The experimental results are shown in Tab.\ref{TAB2}. We have the following observations: CVA obtains the best `all' accuracies on test dev (Open-end 65.92\% vs multiple-choice 70.3\%) and test-standard (66.20\% vs 70.41\%). Our method belongs to the second category (visual attention only). Compared with the second category methods, the improvement over the best approach MCB is significant, by 5.12\% (test-dev, open-ended, all) and 4.9\% (test-dev, multiple-choice, all). In addition, we also report the results of MCB-7 (ensemble of 7 Att. models), although it is not comparable because each model in ensemble of 7 models uses MCB with attention. Comparing with the third category, again our approach outperforms the best approach ACK, especially with an increase of 6.8\% in terms of test-standard opened-end all. Although HieCoAtt and DAN integrate both visual and question attentions, our CVA performs better. For standard-test dataset, tt surpass the DAN by 2.0\% (in open-ended, all) and 1.14\% (in multiple-choice, all), respectively. In addition, compared with MLAN involving both semantic attention and visual attention, our approach is also better, with a clear performance gap on the test-dev dataset. The results in Tab.\ref{TAB2} clearly demonstrate the advantage of our method.

\begin{table}[t]
	\centering
	\scalebox{0.68}{
		\begin{tabular}{|c||ccccc|cc|}
			\hline
			Method     & All   & Obj.  & Num.  & Color & Loc.  & WUPS0.9 & WUPS0.0 \\ \hline \hline
			2VIS+BLSTM & 55.09 & 58.19 & 44.79 & 49.53 & 47.34 & 65.34   & 88.64   \\
			IMG-CNN    & 58.40 & -     & -     & -     & -     & 68.50   & 89.67   \\
			DDPnet     & 61.60 & -     & -     & -     & -     & 70.84   & 90.61   \\
			SAN        & 61.60 & 65.40 & 48.60 & 57.90 & 54.00 & 71.60   & 90.90   \\
			QRU        & 62.50 & 65.06 & 46.90 & 60.50 & 56.99 & 72.58   & 91.62   \\
			HieCoAtt   & 65.40 & 68.00 & \textbf{51.00} & 62.90 & 58.80 & 75.10   & 92.00   \\ \hline
			CVA  & \textbf{67.51} & \textbf{69.55} & 50.76 & \textbf{68.96} & \textbf{59.93} & \textbf{76.70}   & \textbf{92.41}   \\ \hline
		\end{tabular}
	}
	\caption{Evalution results by our proposed method and compared methods on the COCO QA dataset.}
	\label{TAB3}
\end{table}

\begin{table}[t]
	\centering
	\scalebox{0.85}{
		\begin{tabular}{|l|l|l|l|l|l|l|l|}
			\hline
			Methods      & Wht.      & Whr.      & Whn.      & Who       & Why       & How       & Avg \\ \hline
			LSTM-Att     & 51.5      & 57.0      & 75.0      & 59.5      & 55.5      & 49.8      & 54.3         \\ \hline
			MCB+Att      & 60.3      & 70.4      & 79.5      & 69.2      & 58.2      & 51.1      & 62.2         \\ \hline
			MLAN         & 60.5      & \textbf{71.2}      & 79.6      & 69.4      & 58.0      & 50.8      & 62.4         \\ \hline
			\textbf{CVA} & \textbf{64.8} & 59.4 & \textbf{80.1} & \textbf{70.0} & \textbf{65.2} & \textbf{55.7} &    \textbf{63.8}          \\ \hline
		\end{tabular}
	}
	\caption{Evaluation results on Visual7W dataset.}
	\label{tAB3}
\end{table}

\subsection{Comparing with State-of-the-Arts on the COCO-QA dataset}
\textbf{Compared Methods.} In this section, we compare our methods with the state-of-the art approaches on the COCO-QA dataset. We compare it with 2-VIS+BLSTM \cite{nips2015}, IMG-CNN \cite{DBLP:conf/aaai/MaLL16}, DDPnet\cite{DBLP:conf/cvpr/NohSH16}, SAN(2, CNN)\cite{YangHGDS15}, QRU \cite{DBLP:conf/nips/LiJ16} and HieCoAtt \cite{Lu2016Hie}. \eat{Specifically, Ren \textit{et al.} propose to utilize neural networks and visual semantic embeddings to predict answers by directly building on top of the LSTM sentence model. The IMG-CNN applies three convolutional architectures to encode the image content, extract question representations, and to learn one multimodal convolution layers to jointly learn their representation to produce the answer.} In addition, Fig.\ref{fig:2} demonstrates some qualitative results from VQA. 

\textbf{Results.} The experimental results are demonstrated in Tab. \ref{TAB3}. Our CVA achieves the highest performance with an accuracy of 67.51\%  on all and a WUPS 0.9 of 76.7\%. Compared with the best counterpart HioCoAtt and the second best counterpart QRU, the improvement of accuracy is 2.11\% and 5.01\% on all. Specifically, for \textit{Object}, \textit{Color} and  \textit{Loc.} three types questions, CVA increases the accuracy to 69.55\% and 68.96\%, and 59.93\%, respectively. The table indicates that the  \textit{Number} type question performs worst in all methods, with approximately 10 \% to 19\% lower than other types. This might be caused by the unbalanced training dataset. The number of training samples of \textit{Object} type is 54,992, which is nine times more than the number of training samples for \textit{Number}.

\subsection{Comparing with State-of-the-Arts on the Visual7w dataset}
In this section, we further assess our model on the recent released dataset Visual7w. For the experiments, we compare our CVA with previous work LSTM-Att. \cite{DBLP:conf/cvpr/ZhuGBF16}, MCB+Att. \cite{DBLP:conf/emnlp/FukuiPYRDR16} and MLAN  \cite{mutil2017attention} and the experimental results are shown in Tab.\ref{TAB3}. Both  LSTM-Att. and MLAN are competitive. However, compared with LSTM-Att., MLAN utilizes a much lower fusion methods (2400-dimension vs 16,000-dimension)\cite{mutil2017attention}. Tab.\ref{TAB3} shows that our CVA achieves the highest scores. In particular, the MLAN employs both visual attention and text attention, while our CVA only exploits the visual attention.

\vspace{-0.2cm}
\section{Conclusion}
In this paper, we propose a novel cubic visual attention network for visual question answering task. CVA takes the full advantage of characteristics of CNN to obtain visual channel-wise features representing semantic attributes and object region based visual features representing rich semantic information to support visual question answering and it achieves the state-of-the-art on three public standard datasets across various question types, such as multiple-choices and open-ended questions. The contribution of CVA is not only provide a powerful VQA model, but also a better mechanism to understand the visual information for predicting answers. %In addition, both channel-wise and object region based attentions are visual attention, thus in the future we intend to bring question attention in CVA and integrate region relationships to increase VAQ task, especially for the \textit{Number} type question.

\section*{Acknowledgments}
This work is supported by the Fundamental Research Funds for the Central Universities (Grant No. ZYGX2014J063, No. ZYGX2014Z007) and the National Natural Science Foundation of China (Grant No. 61772116, No. 61502080, No. 61632007, No. 61602049).

\bibliographystyle{named}
\bibliography{ijcai18}

\begin{thebibliography}{}

\bibitem[\protect\citeauthoryear{Anderson \bgroup \em et al.\egroup
  }{2017}]{Anderson:2017up-down}
Peter Anderson, Xiaodong He, Chris Buehler, Damien Teney, Mark Johnson, Stephen
  Gould, and Lei Zhang.
\newblock Bottom-up and top-down attention for image captioning and visual
  question answering.
\newblock {\em arXiv preprint arXiv:1707.07998}, 2017.

\bibitem[\protect\citeauthoryear{Antol \bgroup \em et al.\egroup
  }{2015}]{vqa2015}
Stanislaw Antol, Aishwarya Agrawal, Jiasen Lu, Margaret Mitchell, Dhruv Batra,
  C.~Lawrence Zitnick, and Devi Parikh.
\newblock {VQA:} visual question answering.
\newblock In {\em ICCV}, pages 2425--2433, 2015.

\bibitem[\protect\citeauthoryear{Chen \bgroup \em et al.\egroup
  }{2017}]{chen2016sca}
Long Chen, Hanwang Zhang, Jun Xiao, Liqiang Nie, Jian Shao, Wei Liu, and
  Tat-Seng Chua.
\newblock Sca-cnn: Spatial and channel-wise attention in convolutional networks
  for image captioning.
\newblock In {\em CVPR}, 2017.

\bibitem[\protect\citeauthoryear{Fukui \bgroup \em et al.\egroup
  }{2016}]{DBLP:conf/emnlp/FukuiPYRDR16}
Akira Fukui, Dong~Huk Park, Daylen Yang, Anna Rohrbach, Trevor Darrell, and
  Marcus Rohrbach.
\newblock Multimodal compact bilinear pooling for visual question answering and
  visual grounding.
\newblock In {\em EMNLP}, pages 457--468, 2016.

\bibitem[\protect\citeauthoryear{Gao \bgroup \em et al.\egroup
  }{2015}]{gao2015you}
Haoyuan Gao, Junhua Mao, Jie Zhou, Zhiheng Huang, Lei Wang, and Wei Xu.
\newblock Are you talking to a machine? dataset and methods for multilingual
  image question.
\newblock In {\em NIPS}, pages 2296--2304, 2015.

\bibitem[\protect\citeauthoryear{Gao \bgroup \em et al.\egroup
  }{2017}]{GaoGZXS17}
Lianli Gao, Zhao Guo, Hanwang Zhang, Xing Xu, and Heng~Tao Shen.
\newblock Video captioning with attention-based {LSTM} and semantic
  consistency.
\newblock {\em {IEEE} Trans. Multimedia}, 19(9):2045--2055, 2017.

\bibitem[\protect\citeauthoryear{Goyal \bgroup \em et al.\egroup
  }{2017}]{DBLP:conf/cvpr/GoyalKSBP17}
Yash Goyal, Tejas Khot, Douglas Summers{-}Stay, Dhruv Batra, and Devi Parikh.
\newblock Making the {V} in {VQA} matter: Elevating the role of image
  understanding in visual question answering.
\newblock In {\em CVPR}, pages 6325--6334, 2017.

\bibitem[\protect\citeauthoryear{He \bgroup \em et al.\egroup }{2016}]{HeZRS16}
Kaiming He, Xiangyu Zhang, Shaoqing Ren, and Jian Sun.
\newblock Deep residual learning for image recognition.
\newblock In {\em CVPR}, 2016.

\bibitem[\protect\citeauthoryear{Hyeonseob~Nam and Kim}{2017}]{dual2016}
Jung-Woo~Ha Hyeonseob~Nam and Jeonghee Kim.
\newblock Dual attention networks for multimodal reasoning and matching.
\newblock In {\em CVPR}, 2017.

\bibitem[\protect\citeauthoryear{Ilievski \bgroup \em et al.\egroup
  }{2016}]{DBLP:journals/corr/IlievskiYF16}
Ilija Ilievski, Shuicheng Yan, and Jiashi Feng.
\newblock A focused dynamic attention model for visual question answering.
\newblock {\em CoRR}, abs/1604.01485, 2016.

\bibitem[\protect\citeauthoryear{Li and Jia}{2016}]{DBLP:conf/nips/LiJ16}
Ruiyu Li and Jiaya Jia.
\newblock Visual question answering with question representation update
  {(QRU)}.
\newblock In {\em NIPS}, pages 4655--4663, 2016.

\bibitem[\protect\citeauthoryear{Lu \bgroup \em et al.\egroup
  }{2016}]{Lu2016Hie}
Jiasen Lu, Jianwei Yang, Dhruv Batra, and Devi Parikh.
\newblock Hierarchical question-image co-attention for visual question
  answering.
\newblock In {\em NIPS}, 2016.

\bibitem[\protect\citeauthoryear{Lu \bgroup \em et al.\egroup
  }{2018}]{lu2018:co-attending}
Pan Lu, Hongsheng Li, Wei Zhang, Jianyong Wang, and Xiaogang Wang.
\newblock Co-attending free-form regions and detections with multi-modal
  multiplicative feature embedding for visual question answering.
\newblock In {\em AAAI}, 2018.

\bibitem[\protect\citeauthoryear{Ma \bgroup \em et al.\egroup
  }{2016}]{DBLP:conf/aaai/MaLL16}
Lin Ma, Zhengdong Lu, and Hang Li.
\newblock Learning to answer questions from image using convolutional neural
  network.
\newblock In {\em AAAI}, pages 3567--3573, 2016.

\bibitem[\protect\citeauthoryear{Malinowski \bgroup \em et al.\egroup
  }{2015}]{DBLP:conf/iccv/MalinowskiRF15}
Mateusz Malinowski, Marcus Rohrbach, and Mario Fritz.
\newblock Ask your neurons: {A} neural-based approach to answering questions
  about images.
\newblock In {\em ICCV}, pages 1--9, 2015.

\bibitem[\protect\citeauthoryear{Noh \bgroup \em et al.\egroup
  }{2016}]{DBLP:conf/cvpr/NohSH16}
Hyeonwoo Noh, Paul~Hongsuck Seo, and Bohyung Han.
\newblock Image question answering using convolutional neural network with
  dynamic parameter prediction.
\newblock In {\em CVPR}, pages 30--38, 2016.

\bibitem[\protect\citeauthoryear{Ren \bgroup \em et al.\egroup
  }{2015}]{nips2015}
Mengye Ren, Ryan Kiros, and Richard~S. Zemel.
\newblock Exploring models and data for image question answering.
\newblock In {\em NIPS}, pages 2953--2961, 2015.

\bibitem[\protect\citeauthoryear{Ren \bgroup \em et al.\egroup
  }{2017}]{DBLP:journals/pami/RenHG017}
Shaoqing Ren, Kaiming He, Ross~B. Girshick, and Jian Sun.
\newblock Faster {R-CNN:} towards real-time object detection with region
  proposal networks.
\newblock {\em {IEEE} Trans. Pattern Anal. Mach. Intell.}, 39(6):1137--1149,
  2017.

\bibitem[\protect\citeauthoryear{Shih \bgroup \em et al.\egroup
  }{2016}]{DBLP:conf/cvpr/ShihSH16}
Kevin~J. Shih, Saurabh Singh, and Derek Hoiem.
\newblock Where to look: Focus regions for visual question answering.
\newblock In {\em CVPR}, pages 4613--4621, 2016.

\bibitem[\protect\citeauthoryear{Song \bgroup \em et al.\egroup
  }{2017}]{SongGGLZS17}
Jingkuan Song, Lianli Gao, Zhao Guo, Wu~Liu, Dongxiang Zhang, and Heng~Tao
  Shen.
\newblock Hierarchical {LSTM} with adjusted temporal attention for video
  captioning.
\newblock In {\em IJCAI}, pages 2737--2743, 2017.

\bibitem[\protect\citeauthoryear{Wang \bgroup \em et al.\egroup
  }{2017}]{Wang2017fact}
Peng Wang, Qi~Wu, Chunhua Shen, and Anton van~den Hengel.
\newblock The vqa-machine: Learning how to use existing vision algorithms to
  answer new questions.
\newblock In {\em CVPR}, pages 3909--3918, 2017.

\bibitem[\protect\citeauthoryear{Wu \bgroup \em et al.\egroup
  }{2016}]{ask16anything}
Qi~Wu, Peng Wang, Chunhua Shen, Anthony~R. Dick, and Anton van~den Hengel.
\newblock Ask me anything: Free-form visual question answering based on
  knowledge from external sources.
\newblock In {\em CVPR}, pages 4622--4630, 2016.

\bibitem[\protect\citeauthoryear{Xiong \bgroup \em et al.\egroup
  }{2016}]{DBLP:conf/icml/XiongMS16}
Caiming Xiong, Stephen Merity, and Richard Socher.
\newblock Dynamic memory networks for visual and textual question answering.
\newblock In {\em ICML}, pages 2397--2406, 2016.

\bibitem[\protect\citeauthoryear{Xu and Saenko}{2016}]{xu2015ask}
Huijuan Xu and Kate Saenko.
\newblock Ask, attend and answer: Exploring question-guided spatial attention
  for visual question answering.
\newblock In {\em ECCV}, 2016.

\bibitem[\protect\citeauthoryear{Yang \bgroup \em et al.\egroup
  }{2016}]{YangHGDS15}
Zichao Yang, Xiaodong He, Jianfeng Gao, Li~Deng, and Alexander~J. Smola.
\newblock Stacked attention networks for image question answering.
\newblock In {\em CVPR}, 2016.

\bibitem[\protect\citeauthoryear{Yu \bgroup \em et al.\egroup
  }{2017}]{mutil2017attention}
Dongfei Yu, Jianlong Fu, Tao Mei, and Yong Rui.
\newblock Multi-level attention networks for visual question answering.
\newblock In {\em CVPR}, pages 4187--4195, 2017.

\bibitem[\protect\citeauthoryear{Zhu \bgroup \em et al.\egroup
  }{2016}]{DBLP:conf/cvpr/ZhuGBF16}
Yuke Zhu, Oliver Groth, Michael~S. Bernstein, and Li~Fei{-}Fei.
\newblock Visual7w: Grounded question answering in images.
\newblock In {\em CVPR}, pages 4995--5004, 2016.

\end{thebibliography}

\end{document}